\documentclass[oribibl]{llncs}
\usepackage{llncsdoc}

\usepackage[utf8]{inputenc} 
\usepackage[T1]{fontenc}    
\usepackage{hyperref}       
\usepackage{url}            
\usepackage{booktabs}       
\usepackage{amsfonts}       
\usepackage{nicefrac}       
\usepackage{microtype}      
\usepackage{xcolor}
\usepackage{graphicx}
\usepackage{caption}
\usepackage{subcaption}
\usepackage{ulem}
\usepackage{amsmath}
\usepackage{algorithm,algpseudocode}

\title{A Meta-Learning Approach to One-Step Active-Learning}

\author{
  Gabriella Contardo \inst{1} 
  \and 
  Ludovic Denoyer \inst{1} 
  \and 
  Thierry Artieres \inst{2}
}
\institute{Sorbonne Universites, UPMC Univ Paris 06, UMR 7606, LIP6, F-75005, Paris, France. \email{firstname.lastname@lip6.fr} \and Ecole Centrale Marseille-Laboratoire d’Informatique Fondamentale (Aix-Marseille Univ.), France. \email{thierry.artiere@centrale-marseille.fr}}

\begin{document}

\maketitle

\begin{abstract}
We consider the problem of learning when obtaining the training labels is costly, which is usually tackled in the literature using active-learning techniques. These approaches provide strategies to choose the examples to label before or during training. These strategies are usually based on heuristics or even theoretical measures, but are not learned as they are directly used during training. We design a model which aims at \textit{learning active-learning strategies} using a meta-learning setting. More specifically, we consider a pool-based setting, where the system observes all the examples of the dataset of a problem and has to choose the subset of examples to label in a single shot. Experiments show encouraging results.
\end{abstract}

\section{Introduction}

Machine learning, and more specifically deep learning techniques, are now recognized for their ability to obtain high performance on a large variety of problems, from image recognition to natural language processing. However, most of the tasks tackled for now are supervised and need a critical amount of labeled data to be learned properly. Depending on the final application, these labeled examples are often expensive to get (e.g manual annotation), and not always available in large quantity.Learning using a small amount of labeled data is thus a key issue in the machine learning domain.

Humans are able to learn and generalize well from only a few labeled examples (e.g children can recognize rapidly any depiction of a car or some animals -drawing, photo, real life- after having been shown only a few pictures with explicit "supervision"). This problem has been studied in the literature as \textit{one-shot} (or few-shots) learning, where the goal is to predict based on very few supervised examples (e.g one per category). This setting was first proposed in \cite{yip1997sparse}, and it knows a renewal of interest under slightly different flavors. Recently, several methods have been presented, relying on different techniques such as matching networks and bi-LSTM (\cite{vinyals2016matching}) or memory-networks (\cite{santoro2016meta}) which are learned using a meta-learning approach: they aim at learning from a large set of learning problems a strategy that will enable the algorithm to efficiently and rapidly use the (small) supervision when facing a new problem (see Section \ref{sec:rw} for a description of the related work). In this setting, one consider that the model has as an input a set of already labeled data, usually $k$ examples chosen randomly per category in the problem.


In parallel, the field of \textit{active learning} focuses on approaches that allow a model to ask an oracle for the labels of some training examples, to improve its learning. It is thus based on a different assumption where the model has the ability to ask labels for a set of unsupervised data. In this case, different settings can be defined, regarding the nature of the unsupervised examples set (a finite dataset completely observable, i.e \textit{pool-based}, or a stream of inputs), and the nature of the acquisition process (single step or sequential). Some approaches also benefits from an initial small labeled dataset. The decision process for selecting the examples to label being made during training, all methods from state of the art in this field do not \textit{learn} this decision process, but instead design specific heuristics or criterion.  
%
\\~\\
We propose to study a problem at the crossroad of one-shot learning and active learning. We present a method that not only learns to classify examples using small supervision but additionally learns a label acquisition strategy which is used to acquire the training set. We study the case of \textit{pool-based} setting: the model works on a completely observable set of examples. This is novel with regard to previous approach in one-shot learning which consider a stream of examples to classify one after the other. The choice of the subset of examples to label is made in a single step via the acquisition strategy.
\\~\\
In Section \ref{sec2:description}, we define the problem and the specific training strategy inspired from recent one-shot learning methods. We then describe our approach in Section \ref{sec3:modelstatic}, which is based on representation learning and the use of bi-directional recurrent networks. Section \ref{sec4:exp} provides experimental results on artificial and real datasets. 
\section{Related work}
\label{sec:rw}
The active-learning problem has been studied under various flavors, reviewed in a survey in \cite{settles2010}. Generically speaking, methods are usually composed of two components: a \textit{selector}, which decides which examples should be labeled, and a predictor. Most of the approaches focus on sequential labeling strategies, where the system can send some examples to be labeled to the oracle, eventually update its prediction model, and choose new examples to be labeled depending on the answers of the oracle and/or the new predictor. The data examples can be presented to the selector either as a complete set (e.g \textit{pool-based}) or in a sequential fashion, where the selector has to decide at each step if the example should be labeled or not. Several methods for single-instance selector in pool-based setting have been proposed such as \cite{zhang2000value}, which uses Fisher information matrices, or \cite{collet2014optimistic} that relies on a multi-armed bandit approach. Batch-mode (i.e each step can ask for several labels) have been studied for instance by \cite{guo2008discriminative}, using a definition of the performance based on high likelihood of labeled examples and low uncertainty of the unlabeled ones. Stream-based setting have been tackled through measures of "informativeness" (i.e favor labeling of more informative examples \cite{dagan1995committee}), by defining region of uncertainty (e.g \cite{cohn1994improving}), using "committees" for the decision (e.g \cite{Melville2004diverse} with an ensemble method focusing on favoring diversity in committee members). Other types of approaches design decisions by studying the expected model change (\cite{settles2008multiple}) or the expected error reduction (\cite{roy2001toward}). Static methods (i.e where the subset of examples to label is decided in a single shot) have been less studied as it can not benefit from the feedback of the oracle or any estimation w.r.t. the prediction, the quality of the current predictor or  uncertainty measure. However such methods can prove useful when asking several times in a row an oracle is not possible, or when interactions between the learner and the "oracle" is limited, e.g. as cited by \cite{gu2012selective} when using Amazon Mechanical Turks. In this paper, the authors define the problem as \textit{selective labeling}, in a semi-supervised context. They propose to select a subset of examples to label by minimizing the upper-bound of a deterministic out-of-sample error bound for Laplacian regularized Least Squares.  \cite{guillory2009label} present an approach for single batch active learning for specific graphs-based tasks, while \cite{yu2006active} propose a method based on \textit{transductive experimental design}, however they design a sequential optimization algorithm to overcome the combinatorial problem.
\\~\\In parallel, the problem of one-shot learning (first described in \cite{yip1997sparse}) knows a renewal of interest. Notably, recent methods have proposed to use a meta-learning approach, by relying on additionnal data of similar nature (e.g images of different classes). The goal is to design systems that learns to predict on novel problems based only on few labeled examples. For example, \cite{santoro2016meta} propose to use the recent memory-augmented neural network, to integrate and store the new examples. Similarly, \cite{vinyals2016matching} propose to rely on external memories for neural networks, bidirectional LSTM and attention LSTM. One key aspect of their approach is their aim at representing an instance w.r.t. the current memory (i.e observed labeled examples). Note that these approaches design a "one-shot learning problem" (e.g training point/inference point) as a sequential
problem, where one instance arrives after the other. Additionally, the system can receive some afterward feedback on the observed instances.
\\Tackling active-learning through meta-learning has been little studied for now. The work of \cite{woodward2017active} propose an extension of the model of \cite{santoro2016meta}, where the true label of the observed instance is withheld unless the system ask for it. The model can either classify or ask for the label. The decision is learned through reinforcement learning, where the system gets a high reward for accurate prediction and is penalized when acquiring label or giving false prediction. They design the action-value function to learn as a LSTM. This suffers from a similar drawback as one-shot learning methods, as it does not consider the dataset as a whole but instead follow a "myopic" process.
\\The recent work of \cite{bachman2017} is the most closely related to ours, as they propose a similar approach for this novel task of meta-learning an active labeling strategy in a pool-based setting. However, they present a model that sequentially select an item to label in several step, while we propose a "one-step" static selection that does not rely on any oracle feedback.  

\section{Meta-active learning problem and setting}
\label{sec2:description}

\begin{figure}[t!]
\centering
\vspace{-0.5cm}
\includegraphics[width=0.85\linewidth]{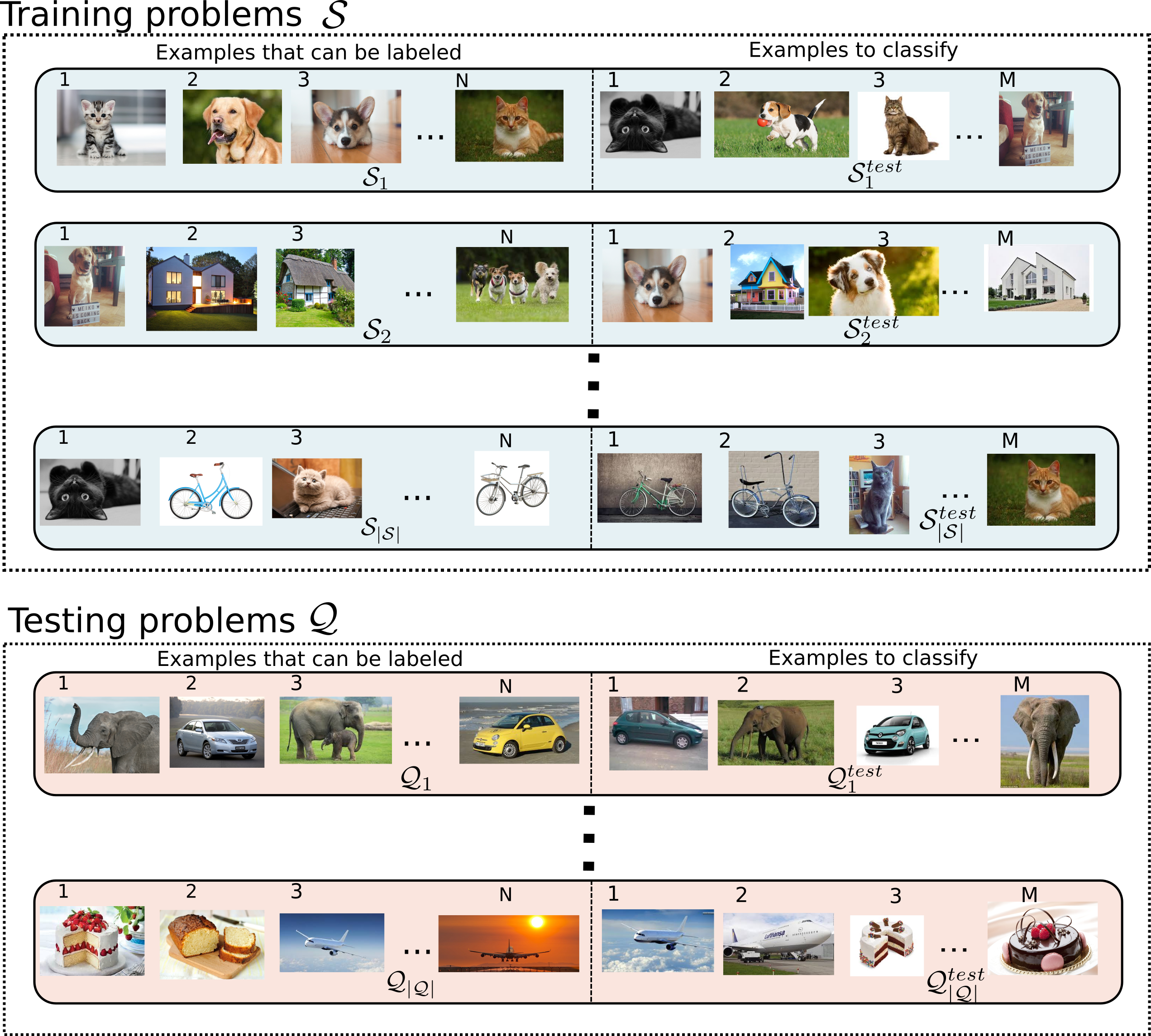}
\caption{Examples of a complete dataset for a meta-active learning strategy, with a set of \textit{training problems} $\mathcal{S}$, with $P$ categories per problem, on a total of $|\mathcal{C}_{train}|$ classes, and a set of \textit{testing problems} on distinct categories. Each problem is composed of a set of $N$ examples that can be labeled and used for prediction, and a set of $M$ examples to classify.}
\label{chap6:fig:metaactive_images}
\end{figure}

\subsection{Preliminary}

The generic goal of an active learning system is to provide the best prediction on a task, using the fewer amount of labels as possible. The system has to choose the most relevant examples to label in order to learn accurately. It is usually considered that the model has access to an oracle, which provides the labels of given examples. Active learning usually aims at tackling a single problem, i.e one dataset and one task. We consider in this paper a pool-based setting with a single-step acquisition, which resumes to the generic following schema: (i) the system receives an entire unsupervised set of examples, (ii) it computes the subset of examples to send to the oracle for labeling, (iii) learning is made based on this reduced supervised subset. In such a single step setting, the decision process for choosing the examples to label can not be learned. 
\\~\\
We propose to design a meta-active learning protocol in order to learn the acquisition strategy i.e the way to choose the examples to label, in a meta learning fashion. We follow a similar principle to what has been recently presented for one-shot learning problems, e.g in \cite{santoro2016meta}. It aims at extending the basic principle of training in machine learning, where a model is trained on data-points from a similar distribution to the data-points observed during inference. For one-shot learning, it resumes as designing data-points as one-shot problems, on dataset of similar nature (e.g all inputs are images). The protocol therefore replicates the final task during training and aims at \textit{learning to learn from few examples}. 
\\
~\\ 
Let us now describe our meta-active learning protocol while introducing few notations. As explained in Figure \ref{chap6:fig:metaactive_images}, our training stage will consist of many elementary active classification problems built from a large dataset. Each elementary problem is denoted $\mathcal{S}=(C, \mathcal{S}^{Train},\mathcal{S}^{Eval})$, it is dedicated to the classification of classes in a set $C$, coming with two sets of examples, the first one being used to infer a prediction model, $\mathcal{S}^{Train}$, and the second one, $\mathcal{S}^{Eval}$,  being used to evaluate the inferred model. 

Starting from a large multiclass dataset $\mathcal{B}$ of labeled examples belonging to a large number of categories $\mathcal{U}^{Train}$, each elementary problem is built as follows:
\begin{itemize}
\item A subset of classes $\mathcal{C}$ is sampled uniformly in the set of all the categories in $\mathcal{U}^{Train}$.
\item Then, a first set of $N$ examples from classes in $\mathcal{C}$ is sampled from $\mathcal{B}$ to build $\mathcal{S}^{Train}=\{ (x_1,y_1),....,(x_N,y_N) \}$, where $x_i$ is the i-th input data-point and $y_i \in \mathcal{C}$ stands for its class. 
\item At last, a second set of $M$ new data points is sampled from $\mathcal{B}$ to build $\mathcal{S}^{Eval}= \{ (x_{N+1},y_{N+1}),....,(x_{N+M},y_{N+M}) \}$ where $\mathcal{S}^{Train} \cap \mathcal{S}^{Eval} = \emptyset$.
\end{itemize}

In the learning stage, the system is presented a series of \textbf{elementary training problems} $\mathcal{S}$. For each problem the training set $\mathcal{S}^{Train}$ is provided without any labels and the system is allowed to ask for the label of a limited subset $\mathcal{D}$ of samples in $\mathcal{S}^{Train}$ according to an acquisition strategy. The system then infers a predictive model $d$ from $\mathcal{D}$ that is evaluated over $\mathcal{S}^{Eval}$. Learning aims at learning the various components of the system (acquisition strategy, infering a predictive model). Each pair $(\mathcal{S}^{Train},\mathcal{S}^{Eval})$ serves as a supervised example for the meta-learning algorithm of the system. 

In the test stage the system is evaluated on \textbf{elementary testing problems}  to evaluate the quality of our meta-learning approach. The testing problems are fully different from training problems since there are  based on a new subset of categories $\mathcal{U}^{Test}$ that is disjoint from the categories used to build the training sets $\mathcal{U}^{Train}$.

An illustration of this setting is provided in Figure \ref{chap6:fig:metaactive_images} with image classification. All elementary classification problems are binary classification (i.e $|\mathcal{C}|$=2). The training problems contains categories such as \textit{cats, dogs, houses} and \textit{bicycles}, with different classification problems e.g classification between \textit{cat} and \textit{dog}, \textit{dog} and \textit{house},etc. The elementary testing problems are drawn from a different set of categories, here \textit{elephants, cars, cakes} and \textit{planes}.

\subsection{Problem Definition}

The goal of a meta-active learning system is to learn an active-learning strategy such that, for each problem, coming with a  training dataset of unlabeled examples, it can predict the most relevant examples to label and provide a good prediction based on these supervised examples on the "test" part of the problem. We propose a system to tackle such a task as composed of two modules. 

The first component is an \textit{active-learning} strategy, which controls the selection of examples. This strategy is defined as a probability distribution over the set of training examples of a problem that we note  $P(\alpha | \mathcal{S}^{train})$ where $\alpha$ is a binary vector of size $N$ such that $\alpha_k=1$ if the strategy asked for label $y_k$ and $\alpha_k=0$ elsewhere. The distribution $P(\alpha | \mathcal{S}^{train})$ is used to sample which examples are asked to be labeled by the oracle. This yields a subset of labeled examples $\mathcal{D}_{\alpha} = \{ x_j \in \mathcal{S}^{train} / \alpha_j=1 \} \subset \mathcal{S}^{train} $. 

The second component is a \textit{prediction} component, which takes as an input an example $x$ to classify in $\mathcal{S}^{eval}$ and the supervised training dataset $\mathcal{D}_{\alpha}$, and outputs prediction for this example denoted $d(x,\mathcal{D}_{\alpha})$. The prediction component does not have access to the examples that have not been targeted by the acquisition policy -- i.e only the examples from $\mathcal{D}_{\alpha}$ are used.

We resume the generic learning scheme in Algorithm \ref{chap6:alglearning}. During training, the process iteratively samples a random problem $\mathcal{S}$ in the set of training problems. The acquisition model receives $\mathcal{S}^{train}$ (without labels) and predicts which examples to select for labelling by sampling with $P(\alpha | \mathcal{S}^{train})$. The built labeled set $\mathcal{D}_{\alpha}$ is used to output prediction for each example in $\mathcal{S}^{Eval}$ using the prediction module $d$. Its performance is evaluated on $\mathcal{S}^{Eval}$ which is used to update the model.
The process is similar at testing time to evaluate the whole meta-learning system.

Since we consider that acquiring labels during the first step has a price, we consider a generic objective function that is a trade-off between the prediction quality on the evaluation set $\mathcal{S}^{Eval}$ and the size of the labeled  set ($|\mathcal{D}_{\alpha}|$), i.e the labeling cost. The generic objective function resumes to:
\begin{equation}
\begin{aligned}
\mathcal{L} &=E_{\mathcal{S} \sim P(\mathcal{S})} [ E_{\alpha \sim P(\alpha | \mathcal{S}^{Train})} [ \sum_{(x_j,y_j) \in \mathcal{S}^{Eval}} [ \Delta(d(x_j,\mathcal{D}_{\alpha}),y_j)] + \lambda |\mathcal{D}_{\alpha}|   ] ]
\end{aligned}
\label{eq:objfunc_generic}
\end{equation}

where $E_{\mathcal{S} \sim P(\mathcal{S})}$ is the expectation over the distribution of problems which we will empirically approximate by an average over a large set of training problems. $E_{\alpha \sim P(\alpha | \mathcal{S}^{Train})}$ stands for the expectation over the subset of examples selected according to the acquisition strategy and  $\Delta (d(x_j,\mathcal{D}_{\alpha}),y_j)$ measures the error $\Delta$ between the expected output $y_j$  and the model prediction $d(x_j,\mathcal{D}_{\alpha})$ for an evaluation sample $x_j$ and a model inferred from $\mathcal{D}_{\alpha}$. 

\begin{figure}[t!]
\centering
\includegraphics[width=0.85\linewidth]{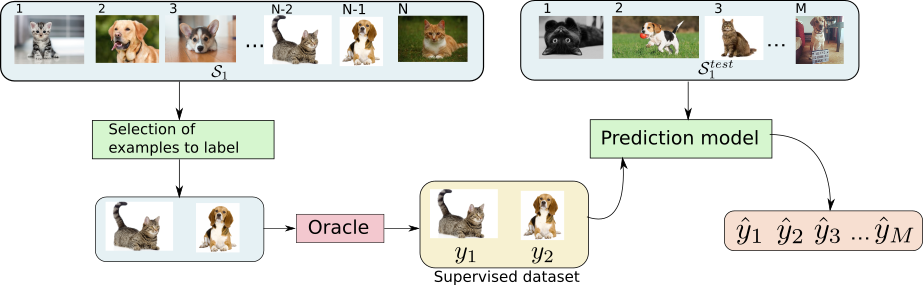}
\caption{Illustration of the inference process for a given problem : the unsupervised dataset $\mathcal{S}_{Train}$ is fed to a "selector", which decides which examples should be labeled. The oracle provides the necessary labels, which provides a small supervised sub-dataset $\mathcal{D}_\alpha$. This dataset is used by the prediction model to predict on the evaluation examples in $\mathcal{S}^{Eval}$.}
\label{fig:active_process}
\end{figure}

 \begin{algorithm}[t!]
\begin{algorithmic}[1]
\Require $ \mathcal{S}$: distribution over training problems.
\Require Active-learning model 
\Require $d$ = Prediction model
  \Repeat
      \State Sample a random problem $\mathcal{S}$ 
      \State Active-learning model predicts the probability $P(\alpha | \mathcal{S}^{Train})$.
      \State Sample following the probability to obtain $\mathcal{D}_{\alpha}$ ,the subset of examples to label in $\mathcal{S}^{Train}$
      \State Feed $d$ with labeled sub-dataset $\mathcal{D}_\alpha$ and evaluate error of $d$ on predictions of all $x_j \in \mathcal{S}^{Eval}$
      \State Update both modules accordingly.
  \Until{stopping criterion}
\end{algorithmic}
 \caption{Learning algorithm for meta-active learning algorithms.}
\label{chap6:alglearning}
 \end{algorithm}

\section{Description of the model}
\label{sec3:modelstatic}
\subsection{Optimization criterion}
We now detail the optimization criterion based on the generic objective function defined in Equation \ref{eq:objfunc_generic}. 
\\
As explained in the previous section, the sub-dataset $\mathcal{D}_\alpha$ of examples chosen for labeling comes from the binary vector $\alpha$, s.t. an example $x_j$ is asked for labeling if $\alpha_j \neq 0$. This vector $\alpha$ is sampled from the distribution $P_\theta(\alpha | \mathcal{S}^{Train})$, outputted by the acquisition component (whose parameters are noted $\theta$), given the unsupervised training set $\mathcal{S}^{train}$. Thus, the number of elements in the dataset $\mathcal{D}_\alpha$ is directly the number of non-zero elements in $\alpha$. The loss for a given problem $\mathcal{S}$ can therefore be rewritten as : 


\begin{equation}
\begin{aligned}
\mathcal{L}_{\theta,d}(\mathcal{S}) &= \mathbb{E}_{\alpha \sim P_\theta(\alpha |\mathcal{S}^{Train})}[\sum_{(x,y) \in \mathcal{S}^{Eval}} {\Delta(d(x,\mathcal{D}_{\alpha}), y) }+ \lambda |\mathcal{D}_{\alpha}|]\\
&= \underbrace{\mathbb{E}_{\alpha \sim P_\theta(\alpha |\mathcal{S}^{Train})}[\sum_{(x,y) \in \mathcal{S}^{Eval}}\Delta(d(x,\mathcal{D}_{\alpha}), y)]}_{\text{error in prediction }} + \underbrace{\mathbb{E}_{\alpha \sim P_\theta(\alpha |\mathcal{S}^{Train})}[\lambda \sum_{k=1}^N \alpha_{k}]}_{\text{cost of labelization} }
\end{aligned}
\label{eq:loss_alleval}
\end{equation}

The first part corresponds to the prediction quality depending on the acquired and labeled examples. Its gradient w.r.t. parameters of both modules (noted for sake of simplicity $\nabla_{\theta,d}$) can be computed using inspired policy-gradient method (likelihood-ratio trick) as follows, where we consider for clarity the gradient of the prediction loss for a single example $(x,y)$ in $S^{Eval}$:

\begin{equation}
\begin{aligned}
\nabla_{\theta,d} \mathbb{E}_{\alpha \sim P_\theta(\alpha |\mathcal{S}^{Train})}[\Delta(d(x,\mathcal{D}_{\alpha}), y)] =& \int \nabla_{\theta,d}(P_\theta(\alpha | \mathcal{S}^{Train}) \Delta(d(x,\mathcal{D}_{\alpha}),y) \mathrm{d}\alpha \\ & + \int P_\theta(\alpha | \mathcal{S}^{Train}) \nabla_{\theta,d} \Delta(d(x,\mathcal{D}_{\alpha}),y) \mathrm{d}\alpha
\\
=& \int \frac{P_\theta(\alpha | \mathcal{S}^{Train})}{P_\theta(\alpha | \mathcal{S}^{Train})} \nabla_{\theta,d} (P_\theta(\alpha | \mathcal{S}^{Train})) \Delta(d(x,\mathcal{D}_{\alpha}), y) \mathrm{d}\alpha \\ &+ \int P_\theta(\alpha | \mathcal{S}^{Train}) \nabla_{\theta,d} \Delta(d(x,\mathcal{D}_{\alpha}),y) \mathrm{d}\alpha
\\
=&\int P_\theta(\alpha | \mathcal{S}^{Train}) \nabla_{\theta,d} (\log(P_\theta(\alpha | \mathcal{S}^{Train}))) \Delta(d(x,\mathcal{D}_{\alpha}), y) \mathrm{d}\alpha \\ &+ \int P_\theta(\alpha | \mathcal{S}^{Train}) \nabla_{\theta,d} \Delta(d(x,\mathcal{D}_{\alpha}),y) \mathrm{d}\alpha
\end{aligned}
\end{equation}
This can be approximated through Monte-Carlo sampling, which yields, on $M$ histories:
\begin{equation}
\begin{aligned}
\nabla_{\theta,d} \mathbb{E}_{\alpha \sim P_\theta(\alpha | \mathcal{S}^{Train})}[\Delta(d(x,\mathcal{D}_{\alpha}), y)] &\approx  \frac{1}{M}\sum_{m=1}^M \nabla_{\theta,d} (\log(P_\theta(\alpha | \mathcal{S}^{Train}))) \Delta(d(x,\mathcal{D}_{\alpha}), y) + \nabla_{\theta,d} \Delta(d(x,\mathcal{D}_{\alpha}),y) 
\end{aligned}
\end{equation}



 \subsection{Labels acquisition component}
 This module  takes as input the whole unlabeled training dataset of the current problem at hand and outputs a probability of the usefulness of labeling each of these samples. We propose to use recurrent neural networks, which were initially proposed to consider sequences of inputs. More specifically, we propose in this work to use \textbf{bi-directional RNN}, which ensure that the output $i$ of the network is computed with regards to all inputs examples, and thus provide a "non-myopic" decision for each example (at the difference of a classical RNN), in order to benefit from the observation of all example for each decision. Note that it could  be relevant to use attentional-LSTM here, as presented in \cite{vinyals2015order}, as it provides an order-invariant network, but this has not been tested yet in our experiments. 
 The output of the recurrent network is  considered to be a probability distribution that is used to sample $\alpha$, the binary vector that select the examples to label. The output can thus be seen either as (i) a multinomial distribution, where $\sum_{i=1}^N \alpha_i=1$, \footnote{Note that this allows to manually bound the number of examples labeled as one has to decide beforehand the number of sampling}, (ii) a bernouilli distribution where each $P_\theta(\alpha_j | \mathcal{S}_i^{train}) \in \{0,1\}$. We present in this paper experiments using a multinomial distribution sampled $k$-times, where $k$ is the maximum number of examples labeled. 
    
    \subsection{Prediction component}
    This module takes as input a (new) example and a limited supervised training dataset, and outputs a prediction (e.g a category). 
 It could be any prediction algorithm, parametric or not, which requires learning or not. In our case, the component should be able to back-propagate some gradients of errors to drive the overall learning. We propose to use similarity based prediction, which doesn't need learning, thus allows for a fast overall meta-learning. We test two similarity measures, a normalized cosine similarity and an euclidean-based similarity. Additionally, computing the predicted label for a new input is done as follow : (i) each similarity with the supervised examples is computed. (ii) This vector of similarities is then converted into a probability distribution, using a softmax with temperature. (iii) The predicted label is computed as the sum of one-hot-vector labels of supervised examples weighted by this distribution. Note that when the temperature is high enough, this distribution is a one-hot vector, which is similar to a 1-nearest neighbor technique.
Additionally, we propose to use a \textbf{representation component}, common to the acquisition and decision components. The key idea is to learn a latent representation space that disentangle the raw inputs to provide better prediction as well as facilitate the acquisition decision. This module, denoted $f$, takes as input an example in $\mathbb{R}^K$ (the original space of all examples of $\mathcal{B}$) and outputs its representation in a latent space $\mathbb{R}^L$. It is conjointly learned with the others functions. Integrating this representation function in the original loss defined in Eq. \ref{eq:loss_alleval} resumes to:
\begin{equation}
\begin{aligned}
\mathcal{L}_{\theta,d,f}(\mathcal{S}) &= \mathbb{E}_{\alpha \sim P_\theta(\alpha f(|\mathcal{S}^{Train}))}[\sum_{(x,y) \in \mathcal{S}^{Eval}}\Delta(d(f(x),f(\mathcal{D}_{\alpha})), y)] + \mathbb{E}_{\alpha \sim P_\theta(\alpha |f(\mathcal{S}^{Train}))}[\lambda \sum_{k=1}^N \alpha_{k}]
\end{aligned}
\end{equation}
Where we note for sake of clarity $f(\mathcal{S}^{Train})=\{f(x_1),\dots,f(x_N)\}$, and similarly for $f(\mathcal{D}_\alpha)$. 

    \section{Experiments}
    \label{sec4:exp}
    We first describe our experimental protocol and the baselines we used, then we show the results of our experiments on two datasets,\textit{letter} and \textit{aloi}.
    
        \paragraph{Experimental Protocol}:
        To build our "meta-active learning" datasets, we set $P$, the number of categories of each elementary problem, $N$ the number of examples in the "unsupervised" dataset, and $M$ the number of examples to classify on. For simplicity, we chose in our experiments  to use same numbers $P, $M$, $N, whatever the every elementary problem. 
        
        The generation of the complete dataset as illustrated in Figure \ref{chap6:fig:metaactive_images} with training/validation/testing problems is based on a partition of the full set of categories between train, validation and test, while keeping a common domain between all inputs. It is done as follows: 
        \begin{itemize}
            \item \textbf{training dataset}: we select a subset of the categories as "training classes" (e.g 50\% of all classes) and their corresponding examples. We then generate a large amount of sub-problems: for one problem, (i) we randomly select $P$ categories in the  "training classes", (ii) we randomly select $N$ examples in these $P$ categories (i.e $\mathcal{S}_i^{train}$, the examples that can be asked for labeling), (iii) we randomly select $M$ additional examples to evaluate the predictions, i.e $\mathcal{S}_i^{eval}$.
            \item \textbf{validation and testing datasets} are generated similarly, on distinct "validation classes" and "testing classes", unobserved in the complete training dataset.
        \end{itemize}

           \paragraph{Baselines}: We propose for this study two baselines. These baselines follow the same global scheme, but with a different acquisition component:
           \begin{itemize}
               \item \textbf{Random }acquisition: the examples to label are chosen randomly in the dataset.
               \item \textbf{K-medoids} acquisition: the examples to label are selected following a k-medoid clustering technique, where we label each example if it is a centroid of a cluster.
           \end{itemize} 
           Note that these acquisition methods do not learn during the overall process, only the representation component (if one is used) is learned. 
           While being simple, we expect the k-medoids baseline to be a reasonable and efficient baseline in our static active-learning setting, more especially when using a similarity-based function for prediction.

\begin{figure}
\vspace{-0.5cm}
 \begin{subfigure}{0.3\textwidth}
  \centering
 \includegraphics[width=\textwidth]{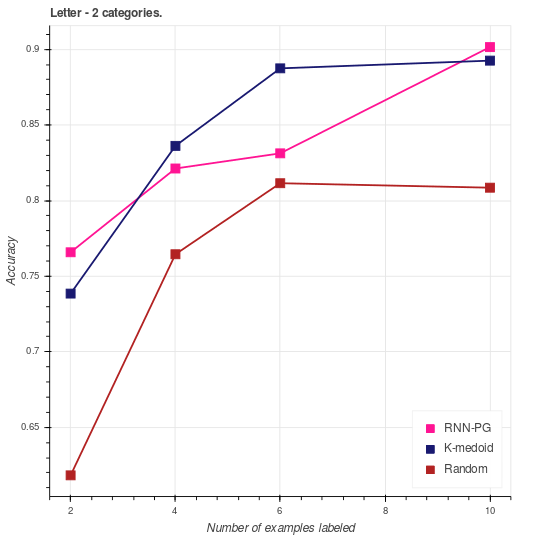}
\caption{Results on dataset \textit{letter} with 2-categories classification problems.}
\label{chap6:letter_2c}
\end{subfigure}
\hspace{0.1cm}
\begin{subfigure}{0.3\textwidth}
  \centering
 \includegraphics[width=\textwidth]{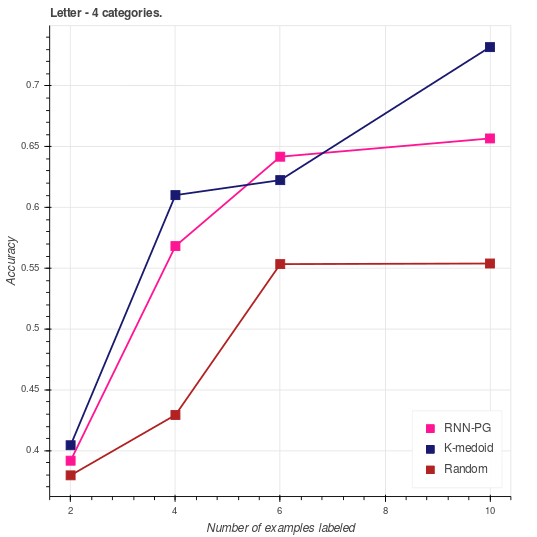}
\caption{Results on dataset \textit{letter} with 4-categories classification problems.}
\label{chap6:letter_4c}
 \end{subfigure}
 \hspace{0.1cm}
\begin{subfigure}{0.3\textwidth}
  \centering
 \includegraphics[width=\textwidth]{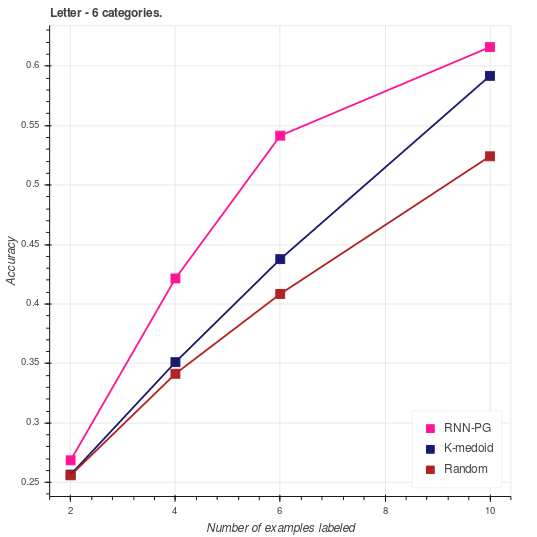}
\caption{Results on dataset \textit{letter} with 6-categories classification problems.}
\label{chap6:letter_6c}
 \end{subfigure}
%
%
 \begin{subfigure}{0.3\textwidth}
  \centering
 \includegraphics[width=\linewidth]{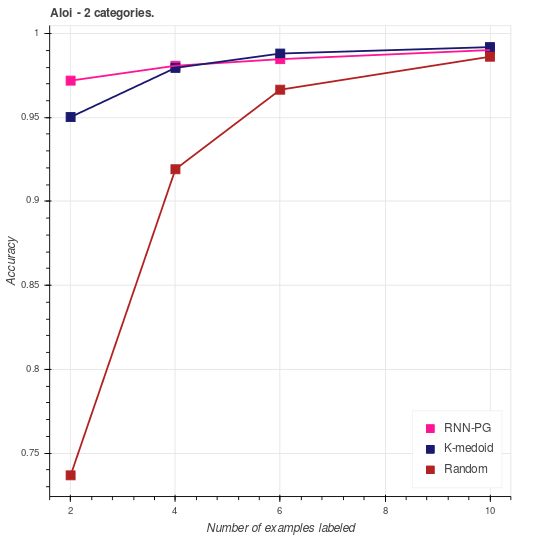}
\caption{Results on dataset \textit{aloi} with 2-categories classification problems.}
\label{chap6:aloi_2c}
\end{subfigure}
\hspace{0.1cm}
\begin{subfigure}{0.3\textwidth}
  \centering
 \includegraphics[width=\linewidth]{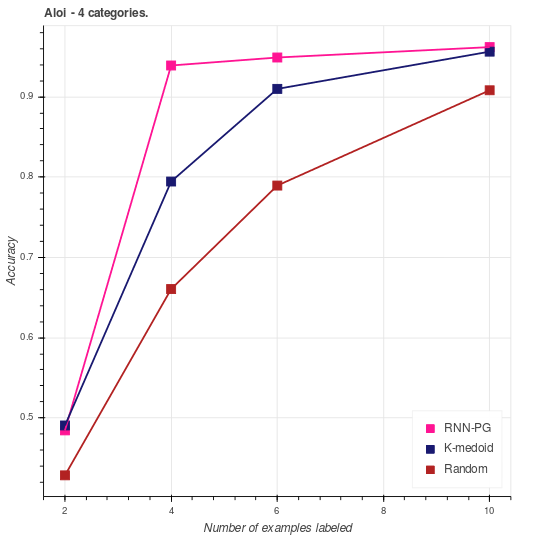}
\caption{Results on dataset \textit{aloi} with 4-categories classification problems.}
\label{chap6:aloi_4c}
 \end{subfigure}
 \hspace{0.1cm}
\begin{subfigure}{0.3\textwidth}
  \centering
 \includegraphics[width=\linewidth]{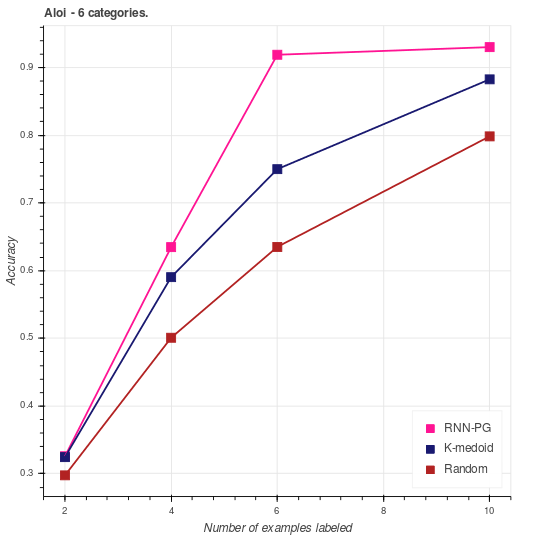}
\caption{Results on dataset \textit{aloi} with 6-categories classification problems.}
\label{chap6:aloi_6c}
 \end{subfigure}
 \caption{Plots of results on uci-dataset \textit{letter} (top) and dataset \textit{aloi} (bottom), with 2,4 or 6 categories per problem. K-medoids acquisition strategy is depicted in blue, random acquisition strategy in red. Our model using Policy-Gradient is in green. Abscissa is the number of examples selected for labeling, ordinate is the average accuracy obtained on all test-problems. For each model, we select the best results on validation problems for each budget, and plot the corresponding performance on test problems (square points).}
 \label{chap6:aloi}
\end{figure}
   
          \paragraph{Dataset letter}: This dataset has 26 categories and 16 features. We took 10 categories for training, 7 for validation and 9 for testing. We generated 2000 problems in training and 500 problems for validation and testing. The size of a dataset (examples that can be labeled) is 25, and the number of examples to classify per problem is 40. Here again we study 3 types of problems, binary, 4-classes and 6-classes with various budget levels. The results are plotted in Figures \ref{chap6:letter_2c},\ref{chap6:letter_4c},\ref{chap6:letter_6c}. We observe mixed results. Our model performs better than a k-medoid acquisition strategy for a budget of 2 on binary-classification problems, but k-medoid leads to a better accuracy for higher budgets. It is also better for all budgets except 6 on 4-categories problems. For 6-categories problems, our model beats the two baselines for all budgets. This difference of performance can be explained by the small amount of different categories in the training dataset; with 10 categories and binary problems (45 different combinations), our model will observe the same problem a large number of times, which could lead to over-fitting. This seems to be the case, as it performs better on 6-classes problems (210 different combinations). We propose thus to study now a dataset with a larger number of categories. 
           \paragraph{Dataset aloi}: This dataset has 1000 categories, with around one hundred images per class. It is a more realistic and challenging dataset for the meta active learning setting we are dealing with. We created 4000 training problems on 350 training categories, and 500 validation and testing problems on respectively 300 and 350 categories. The number of examples that can be labeled is 25, and the number of examples to classify per problem is 40. The results are shown in Figures \ref{chap6:aloi_2c},\ref{chap6:aloi_4c},\ref{chap6:aloi_6c}, for the 3 types of problems (2-classes, 4-classes and 6-classes). We see that our method performs better than k-medoid for all budgets and all types of problems, except on binary-classification with budget 6, where k-medoid performs slightly better (0.5\%). On this bigger dataset, our approach is less prone to overfit, and thus manages to generalize well its acquisition strategy to novel problems on unseen categories.


\section{Closing remarks}
We present in this paper a first approach for a meta-learning approach to a pool-based static active-learning strategy. We propose a stochastic instantiation based on bi-directionnal LSTM to benefit from the whole unsupervised dataset before prediction. First results are encouraging and show the ability of our approach to learn a labeling strategy that performs as well or better than our k-medoid baseline.

\bibliographystyle{splncs03}
   \bibliography{biblio}

\end{document}